# Discovering the Hidden Structure of Complex Dynamic Systems

**Xavier Boyen**
Computer Science Dept.
Stanford University
Stanford, CA 94305-9010
*xb@cs.stanford.edu*

**Nir Friedman**
Inst. of Computer Science
Hebrew University
Jerusalem, 91904, Israel
*nir@cs.huji.ac.il*

**Daphne Koller**
Computer Science Dept.
Stanford University
Stanford, CA 94305-9010
*koller@cs.stanford.edu*

## Abstract

*Dynamic Bayesian networks* provide a compact and natural representation for complex dynamic systems. However, in many cases, there is no expert available from whom a model can be elicited. Learning provides an alternative approach for constructing models of dynamic systems. In this paper, we address some of the crucial computational aspects of learning the *structure* of dynamic systems, particularly those where some relevant variables are partially observed or even entirely unknown. Our approach is based on the *Structural Expectation Maximization (SEM)* algorithm. The main computational cost of the SEM algorithm is the gathering of expected sufficient statistics. We propose a novel approximation scheme that allows these sufficient statistics to be computed efficiently. We also investigate the fundamental problem of discovering the existence of hidden variables without exhaustive and expensive search. Our approach is based on the observation that, in dynamic systems, ignoring a hidden variable typically results in a violation of the Markov property. Thus, our algorithm searches for such violations in the data, and introduces hidden variables to explain them. We provide empirical results showing that the algorithm is able to learn the dynamics of complex systems in a computationally tractable way.

## 1 Introduction

Many real world phenomena are naturally modeled as dynamic systems: the stock market, measurements of a patient's vital signs in an intensive care unit, vehicles on a freeway, etc. Knowledge of a system's dynamics is essential for many tasks, including prediction, monitoring, and the detection of anomalies.

Real world complex systems are typically highly structured, consisting of several interacting entities. *Dynamic Bayesian networks (DBNs)* provide a representational language for modeling such structured systems. By representing a system's state via several variables, and describing the relations between them, a DBN can capture the underlying structure of the system dynamics, e.g., which stocks depend on which other and on what external variables. Compared to less structured representations such as hidden Markov models (HMMs), DBNs support effective approximate inference [Boyen and Koller 1998b] and feature more robust learning due to their reduced number of parameters.

Recent work has made significant progress on the problem of learning Bayesian networks from data (see, for example, [Heckerman 1999]. These ideas have recently been applied to the problem of learning DBNs in the presence of partially observable data [Friedman et al. 1998], using the *Structural EM (SEM)* algorithm [Friedman 1997]. The basic outline of SEM is as follows: Each iteration starts with the current candidate DBN. In the *E-step*, the missing data is completed using the expected value relative to the current DBN, and the completed data is used to gather *expected sufficient statistics* — the completed "counts" corresponding to various events. In the *M-step*, These statistics are then used to score a variety of new candidate structures, and the best scoring candidate is selected. After one or more structural changes, the completed data and statistics are recomputed.

The methods described by Friedman *et al.*, suffer from two major shortcomings. First, their approach has significant computational cost when applied to complex processes such as stock market data. Second, as this algorithm does not change the variables in the network, hidden variables must be prespecified by the user. In this paper, we outline these limitations and suggest methods that address both issues.

The computational cost of the SEM approach is due to the complexity of the E-step. The E-step uses inference to complete the missing data; the inference process propagates messages which are explicit distributions over the set of state variables, so that their representation is exponential in the size of this set. In particular, if a join tree algorithm [Jensen et al. 1990; Shenoy and Shafer 1990] is used for inference, the tree essentially reduces to one huge clique for each pair of consecutive time slices. This exponential cost is prohibitive in all but the simplest DBNs.

This problem also appears in the simpler case of learning DBN parameters given the structure. For that setting, Boyen and Koller [1998a] propose an approximate E-step algorithm. This algorithm propagates approximate messages represented as factorized products over independent clusters. This representation allows the propagation of messages from one time slice to the other using a join tree with much smaller cliques than in the exact method. They show that the error in sufficient statistics resulting from approximation is small, and that the influence on the progress of the learning algorithm is negligible. Even for small networks, order of magnitude speedups can easily be obtained.

In this paper, we extend this technique to the problem of structure learning. In parametric EM, as used in [Boyen and Koller 1998a], each family in the DBN is guaranteed to be contained in some clique in the clique tree for the two consecutive time slices. Thus, sufficient statistics could easily be accumulated during the execution of the inference algorithm. In SEM, on the other hand, the results of the same inference process must be used for scoring a variety of different candidate structures. Each of these requires sufficient statistics for a different set of events. While inference algorithms can in principle be used to compute the probability of any event, this procedure is fairly expensive, especially for a large number of arbitrary events.

In Section 4, we propose a new approximation that circumvents this bottleneck. Roughly speaking, this approximation estimates the posterior probability of the event for which we want statistics by a product of small factors. As we show, this estimate is quite good, and can be computed efficiently as a by-product of the inference algorithm.

Our final contribution addresses a fundamental problem in learning models for dynamic systems. In order to learn models that are statistically robust and computationally tractable, we must often introduce *hidden variables* into our structure. These variables serve many roles: enabling the Markov property, capturing hidden influences of the observables, etc. It is possible, in theory, to discover hidden variables simply by introducing them into the model and hoping that the search algorithm learns their meaning automatically (as in [Friedman et al. 1998]).

We propose a technique that allows for a more targeted search for hidden variables. Our approach is based on the observation that, in DBNs, ignoring a hidden variable typically results in a violation of the Markov property. For example, in a DBN for tracking traffic, eliminating the velocity variable would result in a non-Markovian dependence of the current location on the location at the *two* previous time slices. Our algorithm exploits this property by explicitly searching for violations of the Markov property. For example, a significant correlation between a variable $A^{(t+1)}$ and some other earlier variable $B^{(t-d)}$ (given the Parents of $A^{(t+1)}$) indicates that $A$ must have additional influences not present in our model. In the general case there will be an additional hidden process that influences both $A$ and $B$, and possibly other variables as well. We thus add an extra variable in our model to account for this hidden process.

We believe that our algorithmic ideas combine to provide a viable solution to the problem of learning complex dynamic systems from data. We provide some empirical results on real and synthetic data that, while very preliminary, are quite promising in their ability to scale up to larger systems.

## 2 Preliminaries: Learning DBNs

### 2.1 Dynamic Bayesian Networks

A *dynamic Bayesian network (DBN)* is a model of the evolution of a stochastic system over time. We choose to model the system evolution using a sequence of discrete time points at regular intervals. At each point in time, the instantaneous state of a process is specified in terms of a set of attributes $X_1, \ldots, X_n$. We use $X_i^{(t)}$ to denote the random variable corresponding to the attribute $X_i$ at time $t$.

A DBN represents a distribution over (possibly infinite) trajectories of the system. In order to allow such a distribution to be represented, we usually make two assumptions. The *Markov assumption* states that the future is independent of the past given the present. More precisely, for all $t$, we have the conditional independence $I(\mathbf{X}^{(t+1)}; \{\mathbf{X}^{(0)}, \ldots \mathbf{X}^{(t-1)}\} \mid \mathbf{X}^{(t)})$. This assumption allows us to reduce the representation problem to that of specifying $P(\mathbf{X}^{(0)})$ and $P(\mathbf{X}^{(t+1)} \mid \mathbf{X}^{(t)})$ for all $t$. The second assumption is that the process is *stationary* (or *time-invariant*), i.e., that $P(\mathbf{X}^{(t+1)} \mid \mathbf{X}^{(t)})$ is the same for all $t$.

Given these assumptions, we can specify the proba-bilistic model of a DBN using two *Bayesian network* (BN) fragments. The first is a *prior network* $B_0$ that specifies a distribution over initial states $\mathbf{X}^{(0)}$. The second is a *transition network* $B_{\rightarrow}$, which represents the transition probability from states $\mathbf{X}^{(t)}$ to states $\mathbf{X}^{(t+1)}$. The transition network is a BN fragment over the nodes $\{X_1, \ldots, X_n, X'_1, \ldots, X'_n\}$. A node $X_i$ represents $X_i^{(t)}$ and $X'_i$ represents $X_i^{(t+1)}$. The nodes $X_i$ in the network are forced to be *roots* (i.e., have no parents), and are not associated with conditional probability distributions. We denote the parents of $X'_i$ in the graph by $Pa(X'_i)$. Each node $X'_i$ is associated with a conditional probability distribution (CPD), which specifies $P(X'_i \mid Pa(X'_i))$. The transition probability from one state $\mathbf{x}$ to another $\mathbf{x}'$ — $P(\mathbf{x}' \mid \mathbf{x})$ — is then defined to be $\prod_i P(x'_i \mid \mathbf{u}_i)$, where $\mathbf{u}_i$ is the value in $\mathbf{x}, \mathbf{x}'$ of the variables in $Pa(X'_i)$.

A DBN defines a distribution over infinite trajectories of states. In practice, we reason only about a finite time interval $0, \ldots, T$. To do this reasoning, we can notionally "unroll" the DBN structure into a long BN over $\mathbf{X}^{(0)}, \ldots, \mathbf{X}^{(T)}$. In time slice 0, the parents of $X_i^{(0)}$ and its CPD are those specified in the prior network $B_0$; in slice $t+1$, the parents of $X_i^{(t+1)}$ and its CPD are those specified for $X'_i$ in $B_{\rightarrow}$. Thus, given a DBN $B = (B_0, B_{\rightarrow})$, the joint distribution over $\mathbf{X}^{(0)}, \ldots, \mathbf{X}^{(T)}$ is

$$P_B(\mathbf{x}^{(0)}, \ldots, \mathbf{x}^{(T)}) = P_{B_0}(\mathbf{x}^{(0)}) \prod_{t=0}^{T-1} P_{B_{\rightarrow}}(\mathbf{X}' = \mathbf{x}^{(t+1)} \mid \mathbf{X} = \mathbf{x}^{(t)})$$

## 2.2 Learning DBNs: Complete Data

We now consider the task of learning a DBN from data. For simplicity of notation, we assume that our data $D$ is a single trajectory $\mathbf{d}^{(0)}, \ldots, \mathbf{d}^{(T)}$ through the system; in this section, we assume that this trajectory is fully observable. Also for simplicity, we will ignore the task of learning the prior network $B_0$.

Friedman *et al.* [1998] provide a detailed description of how BN learning can be applied to DBNs. We briefly highlight the most relevant parts of their analysis. Given a training sequence, the learning task is to find the network $B_{\rightarrow}$ that "best matches" $D$. The notion of best match is defined using a scoring function. Several different scoring functions have been proposed in the literature. The most frequently used are the *Bayesian Information Criterion (BIC)* [Schwarz 1978] and the *Bayesian scores*, such as the BDe score of [Heckerman et al. 1995]. Both of these scores combine "fit to data" with some penalty relating to the complexity of the network. For ease of presentation, we will focus on the BIC score.

In both these scores, the term that represents "fit to data" is the *log-likelihood function*, defined as $\ell(B : D) = \log P(D \mid B)$. This function measures the extent to which the data is likely given a candidate model $B$; it is thus an estimate of how well a given candidate fits the empirical data.

The log-likelihood depends on the *sufficient statistics* that summarize the frequencies of the relevant events in the data. For any event $\mathbf{y}$ over $\mathbf{X}, \mathbf{X}'$, we define

$$N_{\mathbf{y}} = \sum_t \iota(\mathbf{y}^{(t)} \mid D)$$

where $\iota(\mathbf{y}^{(t)} \mid D)$ is an indicator function which takes on value 1 if the event $\mathbf{y}$ over $\mathbf{X}, \mathbf{X}'$ holds for the instantiation $\mathbf{X} = \mathbf{d}^{(t)}$ and $\mathbf{X}' = \mathbf{d}^{(t+1)}$, and 0 otherwise. The log-likelihood can now be described as:

$$\ell(B_{\rightarrow} : D) = \sum_i \sum_{x'_i \in \mathrm{Val}(X'_i)} \sum_{\mathbf{u} \in \mathrm{Val}(Pa(X'_i))} N_{x'_i, \mathbf{u}} \log \theta_{x'_i \mid \mathbf{u}} \quad (1)$$

The BIC score is simply the log-likelihood plus a penalty term for network complexity:

$$Score_{BIC}(B_{\rightarrow}) = \ell(B_{\rightarrow} : D) - \frac{\log T}{2} \mathrm{Dim}(G_{\rightarrow})$$

where $\mathrm{Dim}(G_{\rightarrow})$ is the *dimension* of $G_{\rightarrow}$, which in the case of complete data is simply the number of parameters.

Our goal is to find the network that maximizes this score. For a fixed structure, the parameters that maximize the score are exactly the maximum likelihood parameters, which simply mirror the frequencies in the data:

$$\hat{\theta}_{x_i \mid \mathbf{u}} = \frac{N_{x_i \mid \mathbf{u}}}{N_{\mathbf{u}}} \quad (2)$$

Finding the highest scoring network structure is NP-hard [Chickering et al. 1995]. Thus, we usually resort to greedy local search procedures [Buntine 1991; Heckerman et al. 1995] that gradually improve a candidate structure by applying local structural transformation: adding, deleting, or reversing an edge. These transformations are usually applied in a greedy fashion, with occasional random steps to deal with local maxima and plateaux. Two crucial properties of the BIC score greatly facilitate this procedure. First, the score of a network can be written as a sum of terms, where each term determines the score of a particular choice of parents for a particular variable. Thus, a local change to one family $X'_i, \mathrm{Pa}(X'_i)$, such as the addition or removal of an arc, affects only one of these terms. As a consequence, the incremental value of any change to another family in the network remains unchanged. Hence, to determine the values of all local changes to the current network structure we need only re-evaluate changes to the family of $X_i'$.

Second, the term that evaluates the family of $X_i'$ is a function only of the sufficient statistics for $X_i'$ and its parents. Thus, these sufficient statistics are the only aspects of the data that we need to preserve. For each choice of parents for $X_i'$, we need to collect statistics on different events. Evaluation of local changes usually involves computation of new sufficient statistics, and then an evaluation of the score with respect to the statistics of the new model and its dimension.

The Bayesian score is somewhat more complex. It involves taking prior distribution over models and parameters in to account. Without going into details, we note that for some choices of priors, such as the BDe priors of [Heckerman et al. 1995], the main feature of BIC also hold for the Bayesian score: the score decomposes in to a sum of terms, and the score depends only on the sufficient statistics collected from data. Although the Bayesian score and the BIC are asymptotically equivalent, for small sample sizes the Bayesian score often performs better.

### 2.3 Learning DBNs: Incomplete Data

The main difficulty with learning from partial observations is that we no longer know the counts in the data. As a consequence, the score no longer decomposes into separate components corresponding to individual families. The most common solution to the missing data problem is the *Expectation-Maximization (EM)* algorithm [Dempster et al. 1977; Lauritzen 1995]. The algorithm is an iterative procedure that searches for a parameter vector $\boldsymbol{\theta}^*$ which is a local maximum of the likelihood function. It starts with some initial (often random) parameter vector $\boldsymbol{\theta}$. It then repeatedly executes a two-phase procedure. In the E-step, the current parameters are used to *complete* the data by "filling in" unobserved values with their expected values. In the M-step, the completed data is used as if it was real, in a maximum likelihood estimation step.

More precisely, given a current parameter vector $\boldsymbol{\theta}$, the algorithm computes the *expected sufficient statistics (ESS)* for $D$ relative to $\boldsymbol{\theta}$:

$$\begin{aligned} \bar{N}_{\mathbf{y}} &= E[N_{\mathbf{y}} : (G_{\rightarrow}, \boldsymbol{\theta})] \\ &= \sum_t E[\iota(\mathbf{y}^{(t)} \mid D) : (G_{\rightarrow}, \boldsymbol{\theta})] \\ &= \sum_t P(\mathbf{y}^{(t)} \mid D, (G_{\rightarrow}, \boldsymbol{\theta})) \qquad (3) \end{aligned}$$

It then uses the ESS $\bar{N}_\bullet$ in place of the sufficient statistics $N_\bullet$ in (2) for estimating a new set of parameters $\boldsymbol{\theta}$. The process then repeats. The central theorem justifying EM's performance is that each EM cycle is guaranteed to improve the likelihood of the data given the model, until it reaches a local maximum.

EM has been traditionally viewed as a method for adjusting the parameters of a fixed model structure. Friedman's *Structural EM (SEM)* algorithm [1997] extends it to the structure learning task. The SEM algorithm has the same E-step as EM, completing the data by computing expected counts based on the current structure and parameters. In addition to re-estimating parameters, the M-step of SEM uses the ESS, computed according to the current structure, to score other candidate structures. Essentially, the algorithm completes the data using the current network structure, then performs a complete-data structure search in the inner loop. After some number of steps, the algorithm stops the structure search, uses the current candidate network to complete the data, and the process repeats. Friedman [1998] shows that, for a large family of scoring rules, the network resulting from this inner loop must have a higher score than the original. This is true even though the expected counts used in evaluating the new structure are computed using the old structure.

More precisely, Friedman defines a notion of *expected score* which is the expectation of the score for different completions $D^+$ of the data $D$, where the probability of a completion $D^+$ is $P(D^+ \mid D, B)$. For a large class of scores, such as BIC or BDe, he shows that if we make a change to the network structure that increases the expected score, then the true score increases by at least as much. The crucial property of the expected score is that, like the score for complete data, it decomposes into a sum of local terms. For instance, the expected BIC score is simply the BIC score applied to the expected sufficient statistics. This property reinstates both of the important advantages that we had in the case of structure search for complete data: the ability to re-evaluate only a small number of structural changes following a structural change, and the ability to restrict attention to (expected) sufficient statistics.

## 3 The E-step

As the discussion above clearly shows, the key requirement in applying SEM is the ability to compute the expected sufficient statistics. This computation requires that we run probabilistic inference over the entire trajectory $D$, as the distribution over any missing value will typically be affected by past and future evidence. Unfortunately, probabilistic inference for DBNs is notoriously expensive, as Friedman *et al.* [1998] clearly point out in their paper on learning DBNs.

The classical solution to DBN inference is based on the same ideas as the *forward-backward* algorithm for HMMs [Rabiner and Juang 1986].

At a high level, the algorithm propagates *forward messages* $\alpha^{(t)}$ from the start of the sequence forward, gathering evidence along the way; it uses a similar process to propagate *backward messages* $\beta^{(t)}$ in the reverse direction. Letting $r^{(1)}, \ldots, r^{(T)}$ denote the observations along the sequence, $\alpha^{(t)}$ represents the distribution $P(\mathbf{X}^{(t)} \mid r^{(1)}, \ldots, r^{(t)})$; $\beta^{(t)}$ represents the conditional distribution $P(r^{(t+1)}, \ldots, r^{(T)} \mid \mathbf{X}^{(t)})$. The update rules for the messages are:

$$\alpha^{(t+1)}(\mathbf{x}') \propto \sum_{\mathbf{x}} \alpha^{(t)}(\mathbf{x}) P_{B_\rightarrow}(\mathbf{x}' \mid \mathbf{x}) P_{B_\rightarrow}(r^{(t+1)} \mid \mathbf{x}')$$

$$\beta^{(t)}(\mathbf{x}) \propto \sum_{\mathbf{x}'} \beta^{(t+1)}(\mathbf{x}') P_{B_\rightarrow}(r^{(t+1)} \mid \mathbf{x}') P_{B_\rightarrow}(\mathbf{x}' \mid \mathbf{x})$$

Now, the posterior distribution $\phi^{(t)}$ over the states at time $t$ is simply $\alpha^{(t)}(\mathbf{x}) \cdot \beta^{(t)}(\mathbf{x})$ (suitably renormalized). Similarly, the joint posterior $\phi^{(t,t+1)}(\mathbf{x}, \mathbf{x}')$ over the states at $t$ and $t+1$ is proportional to $\alpha^{(t)}(\mathbf{x}) \cdot P_{B_\rightarrow}(\mathbf{x}' \mid \mathbf{x}) \cdot \beta^{(t+1)}(\mathbf{x}') \cdot P_{B_\rightarrow}(r^{(t+1)} \mid \mathbf{x}')$.

This message passing algorithm can be implemented in a straightforward way for DBNs. Indeed, this algorithm was essentially the one used by [Friedman et al. 1998]. Unfortunately, this approach is practical only for very small DBNs: its basic representation — the $\alpha$ and $\beta$ messages — have an entry for every possible state of the system, making them exponential in the number of state variables. Even in highly structured processes, these messages do not admit a more compact representation, as the variables in the message are almost invariably fully correlated [Ghahramani and Jordan 1996a; Boyen and Koller 1998b].

Boyen and Koller [1998b] describe a new approach to approximate inference in dynamic systems, which avoids the problem of explicitly maintaining distributions over large spaces. Their algorithm maintains factored representations that approximate these distributions. In [Boyen and Koller 1998a], they apply this algorithm to the task of message passing in the forward-backward algorithm.

Specifically, they represent messages as products of independent marginals over disjoint clusters. Let $\mathbf{Z}_1, \ldots, \mathbf{Z}_K$ be a partition of $\mathbf{X}$. The approximate message $\tilde{\alpha}^{(t)}$ is represented as a product of marginals $\prod_k \tilde{\alpha}(\mathbf{Z}_k^{(t)})$. Similarly, $\tilde{\beta}^{(t)}$ is represented as $\prod_k \tilde{\beta}(\mathbf{Z}_k^{(t)})$. To apply forward propagation, $\tilde{\alpha}^{(t)}$ is multiplied by the transition model $P_{B_\rightarrow}$, and conditioned on $r^{(t+1)}$. The result is then projected onto each cluster $\mathbf{Z}_k^{(t)}$, and the next message $\tilde{\alpha}^{(t+1)}$ is defined as the product of these marginals. Backward propagation works similarly, except that we take the product of $\tilde{\beta}^{(t+1)}$ by the transition model $P(\mathbf{X}^{(t+1)} \mid \mathbf{X}^{(t)})$, condition on the observations at time $t+1$, and project onto the clusters $\mathbf{Z}_k^{(t)}$ to get $\beta^{(t)}$.

In the case of DBNs, we can accomplish these propagations quite efficiently. For example, in the forward propagation case, we first generate a clique tree $\Upsilon^{(t)}$ [Lauritzen and Spiegelhalter 1988] in which, for every $k$, some clique contains $\mathbf{Z}_k^{(t)}$ and some clique contains $\mathbf{Z}_k^{(t+1)}$. A standard clique tree propagation algorithm can then be used to compute the posterior distribution over every clique. Once that is done, the distribution over $\mathbf{Z}_k^{(t+1)}$ is easily extracted from the appropriate clique potential. (See [Boyen and Koller 1998b] for details.)

Boyen and Koller demonstrate the applicability of their approximate message passing algorithm for the task of parameter estimation for a real-life network with ten state variables (the BAT network of [Forbes et al. 1995]). They show that their algorithm is 10–15 times faster than exact message propagation, and that the degradation in the quality of the learned model is negligible.

## 4 Approximate Sufficient Statistics

This approximate message passing algorithm allows us to perform inference in much larger dynamic systems, and thereby perhaps to learn them. However, we are still faced with a serious problem in computing expected sufficient statistics for structure search. Recall that, to compute ESS, we need to compute joint marginals over sets of variables $\mathbf{Y}$, as in Equation (3). In standard EM for parameter learning, the situation is easy. Here, the only required statistics are the ones over the families of the individual variables in $B_\rightarrow$. By construction of the clique tree, each family is contained in some clique of the clique tree $\Upsilon$ used in our approximate inference. Furthermore, we can use $\Upsilon$ to combine a forward message $\tilde{\alpha}^{(t)}$ and a backward message $\tilde{\beta}^{(t+1)}$, and obtain a compact representation of the full joint posterior $\phi^{(t,t+1)}$ as a tree $\Upsilon^{(t)}$. It is then easy to extract the needed marginals from the appropriate cliques [Boyen and Koller 1998a].

In SEM, this easy solution is no longer applicable, since we use the results of our E-step for a given $B_\rightarrow$ to compute sufficient statistics for a variety of other candidate structures $B'_\rightarrow$. Hence, we will need to compute ESS for events $\mathbf{Y}$ which do not correspond to families in the current structure $B_\rightarrow$. There is no reason to assume that the variables in an arbitrary event $\mathbf{Y}$ will be contained in a single clique in $\Upsilon$. Note that this issue did not arise in the work of [Friedman et al. 1998]. In their approach, the corresponding clique tree to $\Upsilon$ was basically a single clique containing all of the variables in $\mathbf{X}^{(t)}, \mathbf{X}^{(t+1)}$. Thus, all possible queries $\mathbf{Y}$ were contained in a single clique (the one and only clique). It is our ability to provide a finer grained decomposition of the clique tree that brings up this issue.

The straightforward approach to this problem is to compute the necessary posterior $P(\mathbf{Y}^{(t)} \mid D, B_{\rightarrow})$ by performing a special-purpose inference step over $\Upsilon^{(t)}$, tailored to $\mathbf{Y}$. Unfortunately, this operation can be expensive, especially if $\mathbf{Y}$ contains variables which are "distant" in the current structure. It also needs to be performed a great many times every time slice — once for each statistic of interest. The problem is even more serious when we seek statistics over several time slices, as in our search for violations of the Markov property. In this case, $\mathbf{Y}^{(t)}$ contains nodes over several time slices, so that the computation of $P(\mathbf{Y}^{(t)} \mid D, B_{\rightarrow})$ is almost invariably infeasible.

We propose a new approximate solution, in the same spirit as the approximate message decomposition of Boyen and Koller [1998b, 1998a] and of the variational approach of [Ghahramani and Jordan 1996a]. Instead of computing the joint distribution $P(\mathbf{Y}^{(t)})$, we approximate it as a product of independent marginals over individual variables. We approximate $P(\mathbf{Y}^{(t)} \mid D, B_{\rightarrow})$ as $\prod_i P(Y_i^{(t)} \mid D, B_{\rightarrow})$, where for each of the individual variables $Y_i$, $P(Y_i^{(t)} \mid D, B_{\rightarrow})$ is computed by marginalizing some clique in $\Upsilon^{(t)}$ in which the variable is present. From there, $\bar{N}_{\mathbf{Y}}$ is computed as in (3). This process requires a pass over the clique tree to perform the marginalization, and then a simple computation which requires linear time in the number of sufficient statistics that we are maintaining.[1] Furthermore, this approach applies easily to the task of computing ESS for events that span several time slices: we simply extract marginals from more than one clique tree $\Upsilon^{(t)}$. For example, to approximate the joint marginals over an event $X_i^{(t+1)}, X_i^{(t-2)}$, we could extract $P(X_i^{(t+1)})$ from $\Upsilon^{(t)}$ and $P(X_i^{(t-2)})$ from $\Upsilon^{(t-2)}$. We then multiply the two marginals, and add the result to the expected counts for this event, as usual.

At first glance one might think that this approximation discards all correlations between variables in different clusters $\mathbf{Y}_i$. In general, however, this is not the case, since we are examining the posterior distribution $\mathbf{Y}$ given different configuration of evidence. Consider, for example, the situation where we have two binary variables $A, B \in \mathbf{Y}$, which belong to different cliques. If, in our data, we have that, depending on the evidence, $A$ and $B$ are either both probably true or both probably false, then at each step the mass of $(A, B)$ in the product distribution $P(A^{(t)})P(B^{(t)})$ will be large at either $(0,0)$ or $(1,1)$, and its accumulation in our sufficient statistics will reveal the correlation.

[1]We note that we could have used a more refined computation that would have taken advantage of the cooccurrence of some subsets of $\mathbf{Y}$ within a single clique in order to avoid approximating them as independent. However, this extension would require that we marginalize the cliques in a potentially different way for every statistic that we need to compute. Our experiments (see below) suggest that the error introduced by this approximation is probably not large enough to be worth the significant computational overhead.

Table 1: Negative log-likelihood on test data for parametric EM, for different starting points.

| BAT network | Seed #1 | Seed #2 | Seed #3 |
|---|---|---|---|
| Gold standard | 22.1860 | 22.1860 | 22.1860 |
| Exact ESS | 22.4026 | 22.2801 | 22.3269 |
| Approximate ESS | 22.2633 | 22.2676 | 22.2782 |

Of course, there are cases where this approximation *would* lose correlations. For example, if we have that $A^{(t)}$ and $B^{(t)}$ are both uniformly distributed yet correlated, our approximate sufficient statistics would not reveal the correlation. Thus, if the evidence does not give us information about the values of the variables, but only about the correlations, our approximation will lose this correlation. We argue that such models are hard to learn in general (with or without our approximation). Indeed, if the evidence does not give us enough information about the value of a hidden variable, our ability to learn something meaningful about its distribution is very limited.

We tested the error introduced by this approximation on the better-understood problem of parameter estimation. As in [Boyen and Koller 1998a], we generated a long sequence of 20,000 synthetic data points from the BAT network [Forbes et al. 1995]. We then attempted to estimate the parameters back from the data, given the correct structure. We ran two versions of this experiment: one with the approximate message passing algorithm (as above) but without the approximation of the ESS, and the other with both. As can be seen on Table 1, the approximation does not degrade the learning accuracy. On the contrary, the approximation even seems to be slightly beneficial, which could be explained as a regularization effect.

## 5 Structure Search

### 5.1 Efficient Structure Search

In the previous section, we suggested methods for efficient computation of expected sufficient statistics. In SEM search, we need to compute the ESS for each family we change. How do we choose which ESS we should compute?

The first approach is to compute in advance all the expected sufficient statistics the search might need. However, since there are too many of these, this solution is impractical.[2] The second approach, which was used by Friedman et al. [1998], is to compute sufficient statistics "on demand", i.e., statistics for $\mathbf{Y}$ are computed only when the search needs to evaluate a structure with this family.[3] Unfortunately, that also is typically quite expensive, as it requires a traversal over the entire training sequence.

These two solutions are at the extreme ends of a spectrum. Friedman *et al.* [1999] present an intermediate solution which we also adopt. The search procedure works in stages. At the beginning of each stage the search procedure posts the statistics it will require for that stage. These are selected in an informed way, based on the current state of the search. The requested statistics are then computed in one batch, using a single inference phase for all of them at once. More specifically, the algorithm finds for each variable $X'_i$ a set $Pot'_i$ of *potential parents*, based on the current network structure. At each stage, the search is restricted to consider only operations that involve adding edges $Y \to X'_i$ for $Y \in Pot'_i$ or removing current arcs. The number of ESS required for these operations is fairly small, and can be collected at once. The algorithm uses heuristics to focus the attention of the search procedure on "plausible" potential parents. The algorithm therefore requires relatively few statistics in each stage. After this restricted search is done, the process repeats, using the new network as a basis for finding new potential parents for each variable. This process is iterated until convergence of the scoring function.

## 5.2 Discovering Hidden Variables

As we mentioned in the introduction, a fundamental problem when learning dynamic systems from real data is the discovery of hidden variables. In stock market data, for example, internet stocks are typically correlated. Unless we realize that this correlation is due to a hidden variable — public perception of the future growth of internet revenue — the model we learn is likely to be quite poor.

The task of discovering new hidden variables is a notoriously difficult one. In temporal sequences, however, we have some cues that can indicate the presence of such variables. In particular, ignoring a hidden variable will often lead to a non-Markovian correlation, induced by the loss of information as the hidden variable is "forgotten" from step to step. Thus, we can search for non-Markovian correlations, and use them as an indication that the process needs additional "memory" about the past.

More precisely, suppose that we discover that we can predict $X^{(t+1)}$ using $X^{(t)}$ and $Y^{(t-1)}$. Then we might consider creating a new hidden variable $H$ such that $H$ is a parent of $X'$ and $Y$ is a parent of $H'$. Thus, we will have that $Y^{(t-1)}$ influences $H^{(t)}$ via the $Y \to H'$ edge, and that $H^{(t)}$ in turn influences $X^{(t+1)}$ via the $H \to X'$ edge. In other words, $H$ behaves as the "memory" of $Y$ with one step of lag.

In general, we propose the following algorithm. We start by learning the edges among variables in $k$ time slices (a $k$-TBN) for some fixed time window $k$. When some of the variables are unobserved, we use structural EM and our approximation methods to estimate the ESS of the variables in these $k$ consecutive time slices. We note that this process uses structural EM: the sufficient statistics are computed once, and then used for an extended search phase over structures. That, combined with our approach to computing ESS for variables that are far apart in the network, allows us to estimate the ESS for the $k$-TBN without ever doing inference on it.

After we learn such a network, we eliminate the non-Markovian arcs by creating new hidden variables that "remember" those variables that participate as parents in non-Markovian correlations. Any variable $X$ in time slice $t-d$ which directly influences a variable $Y$ at time $t+1$ requires $d$ new hidden variables: at time $t$, the $i$th introduced variable $X^{-i}$ has the same value that $X$ had at time slice $t-i$. In order to represent this "memory" model exactly, the CPDs of these newly created variables would have to be deterministic. However, deterministic models do not easily accommodate EM-style adaptation. Furthermore, since we want to encourage the search to construct variables that remember global phenomena, we also add "persistence" arcs that allow the hidden variables to depend on longer term past. Therefore, we initialize the parameters for these variables to be noisy versions of the appropriate deterministic CPDs, and make the noise biased toward persistence with the previous time slice of the hidden variable.

Having constructed a new 2TBN, we are now again in a position where we can run parametric EM to find better parameters for the new hidden variables. Then

[2] In general, there are exponential number of statistics. When we restrict the indegree of each variable, the number is polynomial, but still unrealistically large.

[3] Of course, we want to avoid unnecessary recomputations. The standard solution is to store a cache of computed statistics, and call the ESS computations on statistics that are not found in the cache.

we repeat the process (structure learning, introduction of variables, etc.).

## 6 Experimental Results

We tried our algorithm on four different domains: three involve real-world data, while the fourth is sampled from a given DBN. We originally ran our algorithm using standard structure search on the observable variables only, allowing for non-Markovian edges. We tested the accuracy on the test set for the learned network, and then used it to introduce hidden variables, as described above. We then repeated this process, running inference to compute a certain predetermined set of ESS, and then using them for an extensive structure search. For structure search, we experimented with both BDe and BIC scores, and with both full and tree-structured CPDs. We report the results for the BDe score with trees; the results for the other cases are somewhat different, but exhibit the same general trends.

As points of comparison, we also learned two other types of structure: a standard Markovian DBN over the observable variables (using a standard BN structure learning algorithm), and a *Factorial HMM (FHMM)* structure [Ghahramani and Jordan 1996b] using parametric EM with the approximate message passing algorithm of [Boyen and Koller 1998a]. An FHMM is best viewed as a DBN with some number $\ell$ of hidden variables, each of which evolves independently of the others; each observable variable depends on all of the hidden variables within its time slice. FHMMs have been shown to be a good candidate for modeling several interacting processes evolving in parallel (e.g., multiple articulatory processes in speech). In our experiments, we tried FHMMs with two, four, and six binary hidden variables. (For data sets with only a small number of observables, we tried fewer hidden variables.)

**Bach chorales** — This data set was proposed as part of the 1991 Santa Fe competition for learning time series [Weigend and Gershenfeld 1990]. It encodes the melodic line of 100 chorales attributed to J.S. Bach. The model has five discrete attributes *Key signature*, *Pitch*, *Duration*, *Fermata*, and *Time signature*. The last three all represent temporal aspects of the piece. The training set consisted of 71 chorales, each of which is about 50 notes long, for a total training set size of 3212 transitions; the test set consisted of 29 chorales, chosen at random, for a total test set size of 1379. The first column of Table 2 shows the results for this data set. We can see that all of the instances of the DBN learning algorithms performed significantly better than all instances of the FHMM algorithm. We also see that the introduction of non-Markovian edges and hidden variables helps significantly with respect to standard structure search over the observables. Somewhat disappointingly, the introduction of hidden variables per se does not improve the log-likelihood. We attribute that to numerical overfitting, as the network structure learned (shown in Figure 1) is actually quite natural. The algorithm detected the correlations between the three tempo attributes. The other two are decoupled from those. All variables have persistence arcs, except (very naturally) the *Fermata* variable, which represents a momentary event corresponding to the end of a segment. The algorithm introduced several hidden variables that capture the non-Markovian nature of these variables. Most interestingly, the *Duration* variable has a time-slice that represents a short non-Markovian dependence (two time slices); on the other hand, the hidden variables introduced for the *Time signature* variable, which represents much longer-term aspects of the piece, correspond to longer-range dependencies.

Table 2: Negative log-likelihood on test data for various algorithms and data sets (in bits/time slice).

| | | Bach | Apnea | Stock | BAT |
|---|---|---|---|---|---|
| Gold standard | | n/a | n/a | n/a | 22.147 |
| Parametric EM | | n/a | n/a | n/a | 22.873 |
| FHMM | 2 hid vars | 8.486 | 3.635 | 24.268 | 23.957 |
| | 4 hid vars | 5.623 | — | 24.302 | 23.562 |
| | 6 hid vars | — | — | 23.213 | 23.773 |
| fully observable only | | 4.538 | 1.892 | 20.834 | 22.693 |
| SEM | iteration 0 | 4.503 | 1.704 | 20.759 | 22.418 |
| | iteration 1 | 4.513 | 1.713 | 20.819 | 22.434 |
| | iteration 2 | 4.537 | 1.710 | 20.710 | 22.388 |

**Sleep apnea** — This data set was also proposed as part of the Santa Fe competition [Weigend and Gershenfeld 1990]. It was obtained by monitoring 3 medical parameters on a patient suffering from sleep apnea. The data contains 34000 points, collected in a single run. This data set is non-stationary, due to the various sleep phases and the condition of the patient. Following the suggestion of Dagum and Galper [1993], each variable was discretized into seven buckets. We partitioned the sequence into four training subsequences, for a total of 19994 transitions, and one test sequence of size 13999. The behavior of the log-likelihoods is very similar to the previous data set. Again, the structure is very natural. There is a strong correlation between *BloodOxygen* and *HeartRate* and between *HeartRate* and *ChestVolume*, but not directly between *BloodOxygen* and *ChestVolume*, a very natural assumption. As above, the algorithm discovered interesting non-Markovian correlations; here, the temporal granularity of *BloodOxygen* is shorter-term than that of *HeartRate*, again, a very reasonable model.

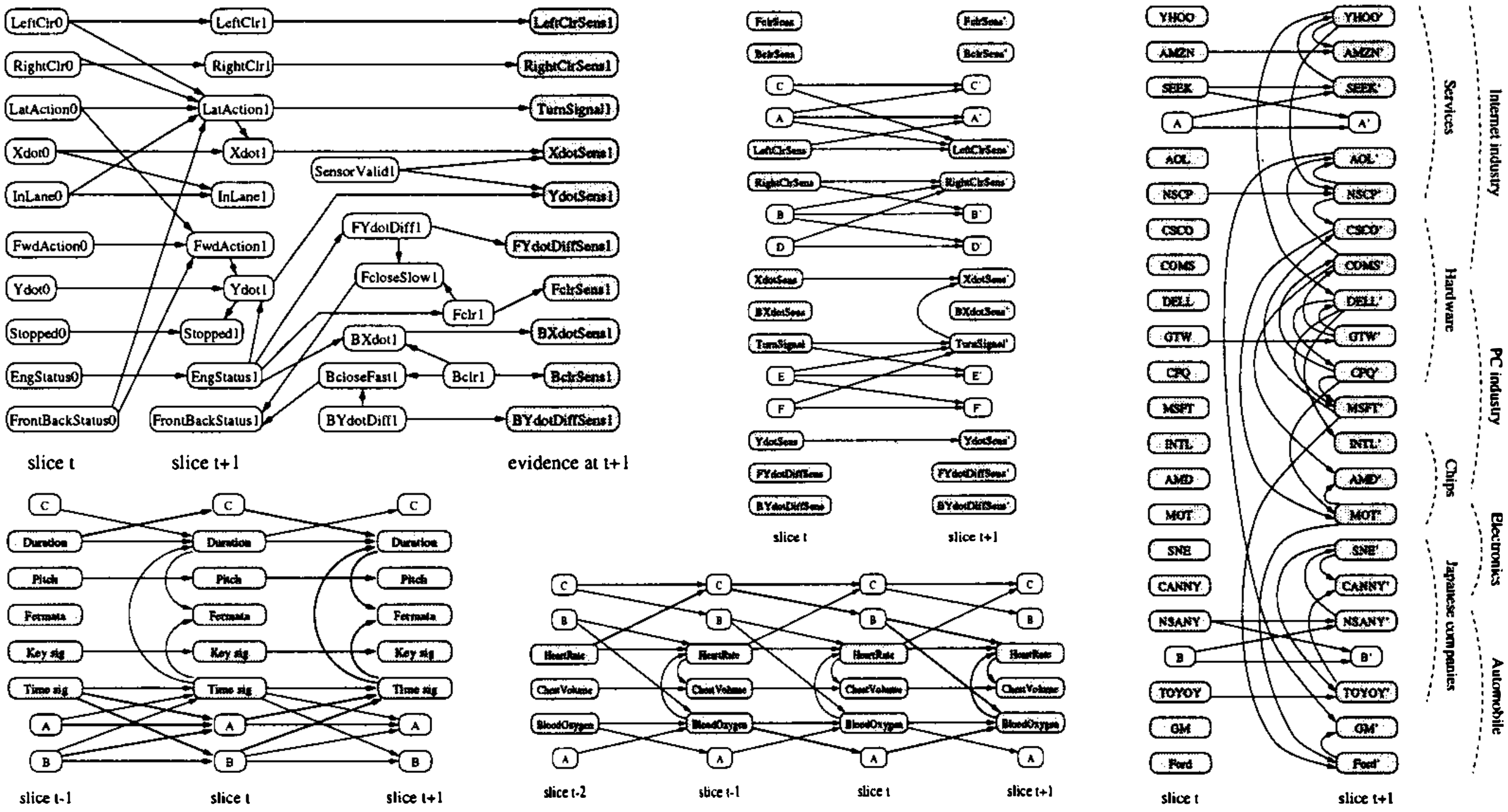

Figure 1: (clockwise from top left) original BAT network; 2TBNs learned from BAT, Stock, Apnea, and Bach. The shaded nodes are observable variables, denoted by their names; hidden variables are labeled alphabetically. (The Apnea and Bach networks have been unrolled for a few time slices to highlight the long-range dependences.)

**Stock market** — We constructed this data set ourselves from the prices of securities of 20 companies in a handful of industries: internet, hardware, software, chips, and car manufacturers from the US and Japan. Since we are usually more interested in trends and correlations than absolute prices, only the daily trend of each stock is recorded (up, down, or n/a). The period covered extends from Feb 92 to Feb 99, for a total of 1768 trading days. We extracted one subsequence from the middle of the period and one from the end for use as training data, giving 1195 transitions for training and 567 for test. The log-likelihood results are shown on Table 2. Here, on the second iteration, the introduction of the hidden variables shows an improvement over the basic score. The shape of the learned network here is somewhat different, as most of the correlations appear within a time slice. This phenomenon is quite natural, as individual trading days can be quite different, but within a given day, many stocks tend to move together. However, as our algorithm is geared to detecting correlations that induce temporal dependencies, it did not discover many hidden variables. To discover hidden variables in this setting, we would need another approach, such as one based on the common intuition of looking for "almost cliques" in the network [Spirtes et al. 1993]. However, we do see that the correlations discovered by our algorithm are quite natural, in that the edges between tend to accumulate between companies in the same industry, or with similar characteristics.

**BAT** — We generated this synthetic data set by sampling a long trajectory from the BAT network [Forbes et al. 1995] for tracking the motion of a car on a freeway. The network has ten state variables, ten observable variables, and a few transient variables (see Figure 1). We extracted four training and five test subsequences, for a total of 4992 training data, and 5029 test data. Only the observable variables were recorded from the trajectory, and the algorithm received no prior knowledge whatsoever about the correct structure. For this domain, since the data is synthetic, we can compute the log-likelihood of the test data according to the correct network, giving us a "gold standard". We see that, in fact, our algorithm learns a model whose performance is fairly close to the gold standard, and much better than parametric EM applied to the correct structure (probably due to numerical overfitting). The performance is also significantly better than the FHMM results or the fully observable structure search. In this case, the introduction of hidden variables actually helps to a nontrivial extent. In this case, however, the learned structures are not very compelling, particularly when compared to the true network. For clarity, we have chosen to present in Figure 1 the somewhat simpler structure learned after only a single iteration of SEM. We can see that the algorithm does discover a few interesting correlations, such as one between *TurnSignal* and *XdotSens* (sensed lateral movement).

## 7 Discussion and Conclusions

In this paper, we combine two lines of works. The first deals with search techniques for learning in the presence of hidden variables [Friedman 1997; Friedman et al. 1998]. The second deals with fast approximate inference in complex networks [Boyen and Koller 1998b; Boyen and Koller 1999]. While approximate DBN inference has been playing a major role in parametric learning [Boyen and Koller 1998a; Ghahramani and Jordan 1996a], this is the first paper to deal with the issues involved in applying it to structure search. In particular, we had to deal with the computation of a large number of different statistics and to introduce methods for discovering hidden variables. Although we based our solution on the Boyen-Koller approximation, many of these ideas can be applied to other approximate inference methods, including the variational methods of Ghahramani and Jordan [1996a].

Clearly, our work only scratches the surface of the problem of discovering hidden variables. While our algorithm discovers correlations that involve temporal interactions, it is less apt at detecting atemporal correlations as we saw in the stock market data. On the other extreme, our algorithm does not support the discovery of truly long-range dependencies and aggregate influences from variables evolving at different speeds. Our current method for discovering hidden influences requires that the time scale of the interaction matches the time scale of the model. If there is a hidden variable evolving much more slowly than the observables, then our algorithm would not find it. This problem can be addressed by explicitly searching for violations of the Markov property at widely varying time granularities. Specifically, applying our algorithm on a data sequence subsampled by a factor of $k$ would exhibit interactions with a time constant of the order of $k$. We believe this issue to be of crucial importance when learning from real-world data. In real systems, observable variables are typically influenced by hidden processes with widely differing time scales, which furthermore are not always related to the sampling rate of the observations.